\documentclass{article}
\usepackage{spconf,amsmath,graphicx}
\usepackage{epsfig}
\usepackage{amssymb}
\usepackage{subfig}
\usepackage{cite}
\usepackage{multirow}
\usepackage{epstopdf}
\usepackage{graphicx}
\usepackage{amsmath}
\usepackage{color, soul}
\usepackage[style=base]{caption}
\usepackage{lineno}
\usepackage{comment}
\usepackage{array}  
\usepackage{siunitx}
\usepackage{mathtools}
\usepackage{calrsfs}
\usepackage{eucal}
\usepackage{bm}
\newcolumntype{d}{>{\displaystyle}c}
\usepackage{tabularx}

\usepackage{subfig}

\usepackage{titlesec}

\titleformat{\paragraph}
{\normalfont\normalsize\bfseries}{\theparagraph}{1em}{}
\titlespacing*{\paragraph}
{0pt}{3.25ex plus 1ex minus .2ex}{1.5ex plus .2ex}

\usepackage{url}
\usepackage{hyperref}
\hypersetup{
	colorlinks=true,       % false: boxed links; true: colored links
	linkcolor=red,          % color of internal links (change box color with linkbordercolor)
	citecolor=green,        % color of links to bibliography
	filecolor=magenta,      % color of file links
	urlcolor=cyan           % color of external links
}
\usepackage{enumitem}
\usepackage{pgfplots}
\usepackage{array}
\newcolumntype{L}[1]{>{\raggedright\let\newline\\\arraybackslash\hspace{0pt}}m{#1}}
\newcolumntype{C}[1]{>{\centering\let\newline\\\arraybackslash\hspace{0pt}}m{#1}}
\newcolumntype{R}[1]{>{\raggedleft\let\newline\\\arraybackslash\hspace{0pt}}m{#1}}
\title{BOTTOM-UP APPROACHES FOR MULTI-PERSON POSE ESTIMATION and it's Applications: A brief Review}
\name{
	\begin{tabular}{ccc}
		 Milan Kresovi\'c & Thong Duy Nguyen
	\end{tabular}
}

\address{
	\begin{tabular}{c}
		Norwegian University of Science and Technology, Norway.
	\end{tabular}
}

\begin{document}
%\ninept
%
\maketitle

\begin{abstract}
Human Pose Estimation (HPE) is one of the fundamental problems in computer vision. It has applications ranging from virtual reality, human behavior analysis, video surveillance, anomaly detection, self-driving to medical assistance.  The main objective of HPE is to obtain the person's posture from the given input. Among different paradigms for HPE, one paradigm is called bottom-up multi-person pose estimation. In the bottom-up approach, initially, all the key points of the targets are detected, and later in the optimization stage, the detected key points are associated with the corresponding targets. This review paper discussed the recent advancements in bottom-up approaches for the HPE and listed the possible high-quality datasets used to train the models. Additionally, a discussion of the prominent bottom-up approaches and their quantitative results on the standard performance matrices are given. Finally, the limitations of the existing methods are highlighted, and guidelines of the future research directions are given.  
\end{abstract}

\begin{keywords}
Human pose estimation, 2D and 3D bottom-up approaches, Multi-person pose estimation
\end{keywords}
%

%------------------------------------------------------------------------------------------------------------------%

\section{Introduction:}

\begin{figure}
	\centering
	\includegraphics[scale=1]{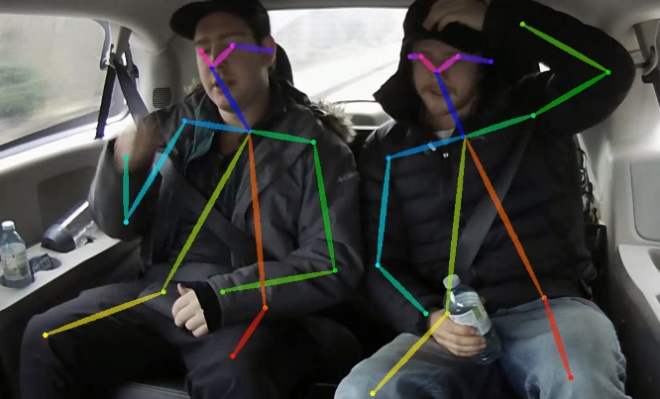}
	\caption{Example of the output of HPE where the key joints of two person sitting in a car are detected.} 
	\label{fig1}
\end{figure}

Human pose estimation (HPE) is an active field of research, and it received enormous interest from different research communities due to its usefulness and versatility. The human pose estimation problem can simply be deduced as obtaining the person's posture from the given input. From a technical perspective, the general overview of the problem that HPE methods are trying to solve is the following:
\begin{itemize}
	\item Find the best method for obtaining the location of the human body key points 
	\item Find the best method for grouping localized key points into a valid human pose configuration 
\end{itemize}
HPE is considered relatively low level computer vision and machine learning tasks. However, it's importance comes from the fact that it plays a crucial role in optimizing the workflow of other important applications domain such as virtual reality \cite{schmidtke2021unsupervised, ullah2021social, chen2021sgpa}, crowd analysis \cite{ullah2021multi,  choi20213dcrowdnet, 9,riaz2021anomalous, stenum2021two, 56}, animal farming \cite{fang2021pose, 54}, crowd coherency detection \cite{japar2021coherent, 66, 11, ullah2016, 12, CrowdCoherency}, intelligent visual surveillance \cite{10, 55}, action recognition \cite{ullah2021attention, moon2021integralaction, ullah2019two, wu2021pose, kanwal2019image, ullah2017human, cai2021jolo, ullah2019stacked}, segmentation \cite{zhuang2021semantic, 55, zhai2021jd, 68, ullah2018pednet}, object detection \cite{hofer2021object, PedAppearance, chen2021monorun, 64, varamesh2020mixture, 67, hagelskjaer2020pointvotenet, 51}, autonomous driving \cite{zhao2021shape, ding2021globally, gu2019efficient} tracking \cite{shagdar2021geometric, chen2020cross, 65, wang2020combining, 58, zhou2020temporal, 59, ariz2016novel, 57, ariz2016novel, 60, chu2021part, 61, zhang2021voxeltrack, 53}, medical imaging \cite{achilles2016patient, mcnally2020evopose2d, bala2020automated, casas2019patient} and facial emotion recognition \cite{bisogni2021ifepe, 62} to name to a few. In a nutshell, HPE methods try to detect and determine the spatial location of body keypoints, i.e. parts or joints, from the given input. Later, from these keypoints, obtain the pose of the human body \cite{munea2020progress}. Figure \ref{fig1} shows an example of the output of an HPE Model. 
This review paper aims to give an overview of recent 2D and 3D multi-person HPE methods based on bottom-up approaches. Different algorithms are compared based on their characteristics, evaluation on different datasets, and inference latency. In the following subsections, the classification of HPE methods is elaborated based on dimensions, the number of objects to estimate the pose for, and possible paradigm. 
\begin{figure}[!htp]
	\centering
	\includegraphics[scale=0.44]{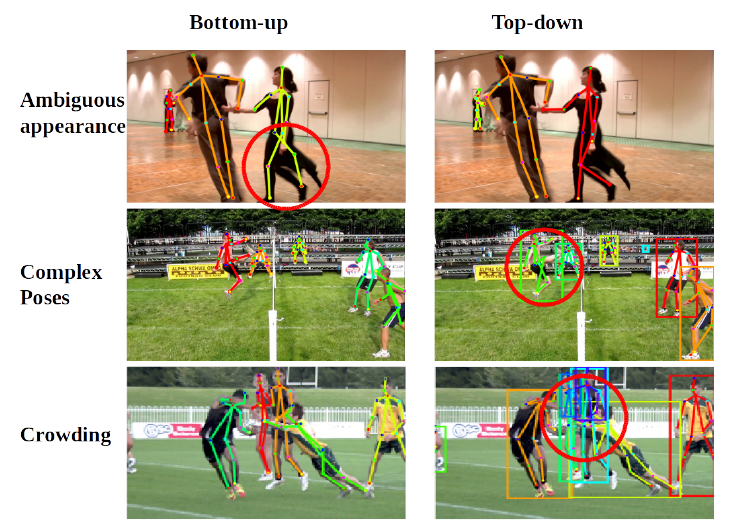}
	\caption{Comparison of the caveats of bottom-up and top-down approaches \cite{jin2017towards}} 
	\label{fig2}
\end{figure}

\subsection{2D Vs. 3D pose estimation}
HPE can be done either in 2D or 3D. 2D HPE represents obtaining the 2D coordinates of anatomical keypoints using monocular images or videos. In the earlier works, this was usually done using pictorial structures \cite{andriluka2009pictorial}. Pictorial structures can be described as tree-structured graphical models that show spatial correlations of the body parts. Although these structures were successful in estimating human pose in images that don’t have too many occlusions, when that is not the case, this approach shows poor results \cite{andriluka2009pictorial}. Another type of approach was using hand-crafted features. This kind of approach was before deep learning models started being more widely used. It was heavily dependent on human supervision. Some of those hand-crafted features were HOGs, color histograms, edges, contours, etc. \cite{yang2011articulated}. Although these kinds of features can easily be obtained, they lack generalization and often produce bad results. On the other hand, the 3D HPE objective is to get the coordinates of anatomical key points in 3D space using monocular input or Multiview cameras. This type of HPE is often categorized as model-free and model-based, depending on whether the human body model is used \cite{sarafianos20163d}.

\subsection{Single Vs. Multi-person HPE}
Depending on the number of present persons on a given input, HPE approaches can be divided into a single person and a multi-person approach. Single-person HPE is used for estimating the human pose for just one person in the image. That is, it is guaranteed that the input contains only one person. As this is not the case in most cases, another type of approach is needed to be developed. If there is more than one person in the given input, then that kind of approach is called multi-person pose estimation. For example, for the self-driving car to be safer with pedestrians or for video surveillance to have better tracking, there is a massive necessity for obtaining a fast and reliable multi-person HPE algorithm.

\subsection{Top-down Vs. Bottom-up}
HPE methods can be characterized as top-down or bottom-up approaches. Top-down approaches first detect all persons from the input and later apply a human pose estimation algorithm for each detected person independently. On the other hand, bottom-up approaches first detect all body parts from the input and later group them for each person. There are several caveats to these two approaches. Top-down approaches have poorer performance in terms of time. These methods are usually time-consuming and computationally complex. Additionally, the performance of top-down approaches is heavily influenced by the performance of the technique used for person detection. Therefore, top-down approaches have a problem when there are complex poses and when there is a lot of crowding. On the other hand, even though bottom-up approaches are faster, they have some issues, such as detecting and grouping body parts when there is a crowd of people or when there is an ambiguous appearance of the body parts in the input. Figure \ref{fig2} shows a comparison of these two approaches.

\subsection{Contributions}
Currently, the problem of HPE is an active field of research. Since the last decade,  tremendous progress has been made in human pose estimation and the datasets that have been created.  This paper provides an overview of the most recent HPE methods and points to future research directions like some surveys as mentioned earlier. The concept that differentiates this survey from the other surveys is that it focuses only on bottom-up multi-person pose estimation. Here is worth noting that a review paper has not yet been released with a focus on that.  These types of approaches are the most promising for real-time implementation and, as such, can serve a great deal for a variety of industries. In summary, the contribution of this survey can be condensed into three aspects:
\begin{itemize}
	\item Provides an overview of the most recent 2D and 3D multi-person HPE methods that are based on bottom-up approaches 
	\item Categorizations of presented HPE methods 
	\item Presentation of the limitations and open issues for each method 
\end{itemize}

This review paper is organized as follows. Section \ref{sec:2} introduces and describes some of the ways to categorize HPE methods.  Section \ref{sec:3} describes the existing datasets that are widely used in HPE evaluation and presents some of the performance metrics. Section \ref{sec:4} gives an overview of some of the state-of-the-art methods, both 2D and 3D. Discussion about the presented methods, existing issues, and several promising future directions is presented in section \ref{sec:5}. Last but not the least, section \ref{sec:6} concludes the paper.

\begin{figure*}[!htp]
	\centering
	\subfloat[\centering]{{\includegraphics[width=7cm]{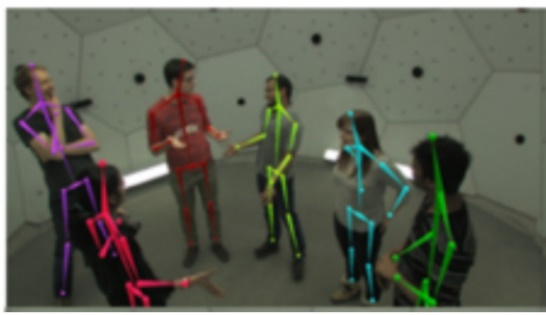} }}%
	\qquad
	\subfloat[\centering]{{\includegraphics[width=7cm]{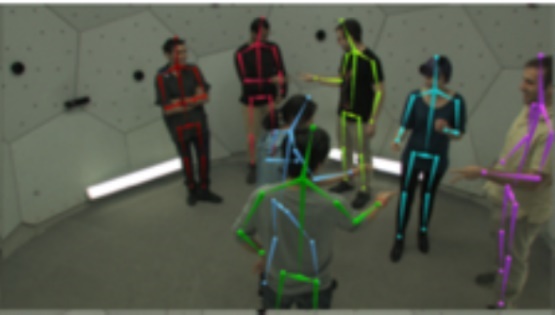} }}%
	\caption{Examples from panoptic dataset}%
	\label{fig4}%
\end{figure*}
\section{Related Work}
\label{sec:2}
In the last couple of years, there has been a couple of HPE review papers, specifically for 3D HPE (\cite{chen2020monocular}, \cite{sminchisescu20083d}), model-based HPE (\cite{chen2020monocular}, \cite{holte2012human}), monocular-based HPE (\cite{chen2020monocular}, \cite{sminchisescu20083d}), etc.  Some of the review papers are just focused on methods that don’t implement deep-learning approaches (\cite{gong2016human, zhang2016survey}), while others \cite{guo2016deep} focus on methods based on deep-learning. A comprehensive review of the 2D and 3D HPE is presented in \cite{chen2020monocular}. In this paper, 2D HPE is categorized like it is in the \cite{dang2019deep}, which is a review paper oriented only at 2D HPE approaches. On the other hand, in \cite{chen2020monocular}, 3D HPE methods are categorized as in \cite{sarafianos20163d}. This review paper focuses on 3D HPEs and categorization of these HPEs into three distinctive groups:
\begin{itemize}
	\item Generative vs. Discriminative methods 
	\item Regression-based vs. Detection-based methods 
	\item One-stage vs. Multi-stage methods 
\end{itemize}

\subsection{Generative vs. Discriminative Models}
Generative and discriminative models mainly differ in whether a particular method uses the human body model or not. 
Generative models take the body model as a priori information. In a more general form, these methods can be deconstructed into two phases \cite{sminchisescu2002estimation}. First is the modeling phase, where a likelihood function is constructed by taking into account the camera model, the human body model, all constraints, etc. After constructing this likelihood function, the second phase is introduced, and that is the estimation phase. In this phase, the human body poses are predicted based on given input and the constructed likelihood function \cite{sarafianos20163d}. Besides this general form of generative methods, there is also a subcategory of generative approaches that is called part-based. Instead of having a rigid human body model, in these methods, the human skeleton is represented as a collection of body parts. These parts are only constrained by the joints in the body model. This allows for a collection of the body part that can take a deformable configuration \cite{sarafianos20163d}.
Discriminative models don't take into account any specific human body model. These methods learn to map given input directly to the human pose space or to search for example pool of pose descriptors. Discriminative methods are in most cases faster than generative methods, but they lack the robustness for the poses that were not introduced during the training \cite{chen2020monocular}.
\begin{figure*}[!htp]
	\centering
	\subfloat[\centering COCO keypoints \cite{lin2014microsoft}]{{\includegraphics[width=5cm]{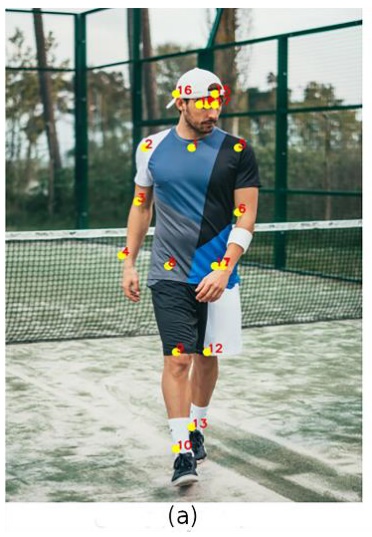} }}%
	\qquad
	\subfloat[\centering MPII keypoints \cite{andriluka20142d}]{{\includegraphics[width=5cm]{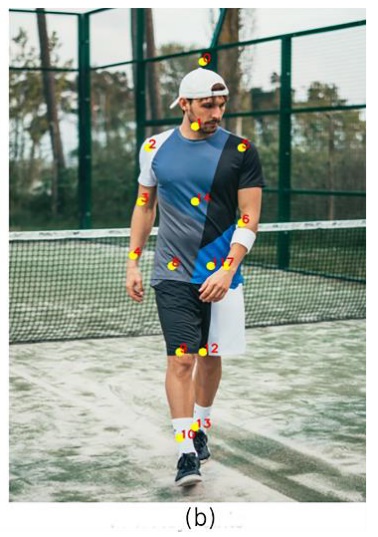} }}%
	\caption{Human keypoint location based on COCO and MPII format}%
	\label{fig3}%
\end{figure*}
\subsection{Regression Vs. Detection}
Discriminative HPE methods for the most part represent methods that are based on deep learning. These methods can further be differentiated based on the way that they detect body joints. When discriminative methods directly map given input to the coordinates of body joints, then these methods can be categorized as regression-based (also known as learning-based) approaches. For this to function method needs to generalize well for a new set of images from the testing set \cite{huang2009estimating}. Detection-based approaches look at the body parts as the detection targets. These targets are usually represented by two different representations \cite{chen2020monocular}:
\begin{itemize}
	\item Image patches 
	\item Heatmaps of joint locations 
\end{itemize}
Additionally, if the discriminative method utilizes an example pool of the pose descriptors, then that method is categorized as an example-based approach.  The final pose is calculated by doing the interpolation of the candidates that are obtained with the similarity search \cite{sarafianos20163d}. Because of the high non-linearity of the problem that regression-based approaches try to solve, detection-based approaches are more widely used. Although, the disadvantages of these types of approaches is that compared to the original images, newly created representations have a smaller resolution, therefore the accuracy of the predicted location of the body joint is smaller \cite{chen2020monocular}.

\subsection{ONE-STAGE VS. MULTI-STAGE}
The main difference between one-stage approaches and multi-stage approaches is whether the end-to-end network is used. That is, multi-stage methods have two or more different stages that are specialized for each task. On the other hand, one-stage methods employ only one end-to-end network to map given input to human poses \cite{chen2020monocular}.
One of the examples for multi-stage methods can be seen in 3D HPE. Some of the methods first predict body part locations on the 2D surface, and later extend that 2D surface to the 3D space.

\section{Datasets}
\label{sec:3}
Datasets are an important component of any HPE method. Not only are they important for evaluating the method, but more complex datasets give better agility for researchers to try out new ideas. Making HPE datasets is a laborious and long process. In the past, the first datasets didn’t have a lot of images, and the complexity was low. This is due to the fact that the process Is time-consuming and finding the people willing to do it was harder. As crowdsourcing services began to rise, datasets became better and no longer bounded by the lab environments or the data quantity.This section will differentiate existing datasets whether they are 2D or 3D. Additionally, this section will only focus on datasets that can be used for multi-pose HPE. At the end of this section, some of the performance metrics will be presented.

\subsection{2D datasets}
Two biggest and most used datasets for 2D multi-person HPE are:
\begin{itemize}
	\item Max Planck Institute for Informatics (MPII) Human Pose Dataset \cite{andriluka20142d} 
	\item Microsoft Common Objects in Context (COCO) Dataset \cite{lin2014microsoft} 
\end{itemize}
\begin{figure*}[!htp]
	\centering
	\includegraphics[scale=0.9]{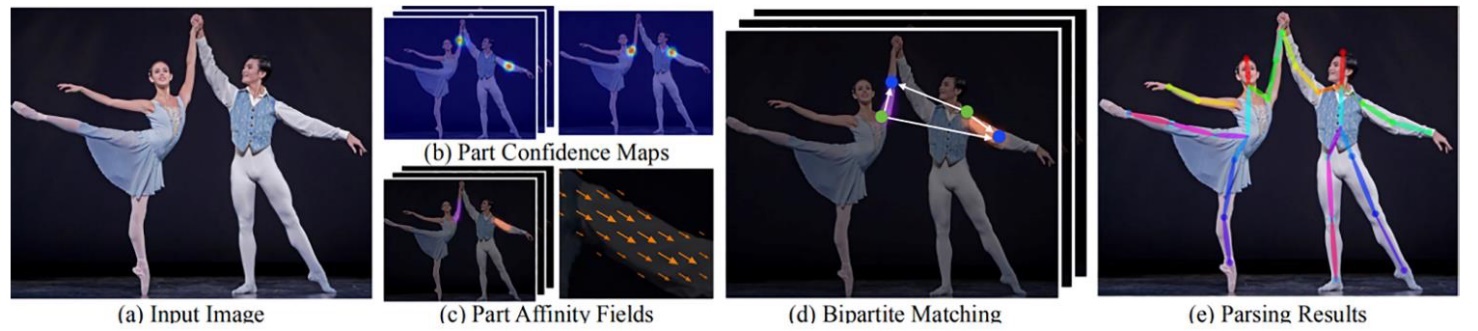}
	\caption{The workflow of the OpenPose model \cite{cao2017realtime}} 
	\label{fig5}
\end{figure*}
\subsubsection{MPII DATASET}
MPII Human pose dataset is one of the state-of-the-art benchmarks for evaluation of articulated human pose estimation \cite{andriluka14cvpr}. In order to produce this dataset 24,920 frames were selected from 3,913 YouTube videos that contain either different people in the same setting or the same person in different poses. The frames are annotated by 15 body joints and a background keypoint, in addition to the 3D viewpoint of the head and torso. Besides the position of each keypoint, this dataset also contains information about the visibility of each keypoint.

\subsubsection{MS-COCO DATASET}
MS-COCO dataset is widely used for multi-person HPE methods. It is a dataset that contains object detection, segmentation, image captioning, and keypoint detections. Images were collected from most popular search engines like Google, Bing, etc. The dataset contains 200,000 images and around 250,000 labeled person instances \cite{chen2015microsoft}. Each person is annotated by 18 keypoints and one for the background. Like the MPII dataset, besides the position of each keypoint, this dataset stores the information about the visibility of each keypoint. Figure \ref{fig3} shows the comparison of COCO and MPII keypoints.
\subsection{3D datasets}
Mostly used datasets for 3D multi-person HPE methods are the following:
\begin{itemize}
	\item Panoptic dataset  \cite{joo2017panoptic} 
	\item 3DPW dataset \cite{huang2009estimating}
\end{itemize}
\begin{figure*}[!htp]
	\centering
	\includegraphics[scale=0.4]{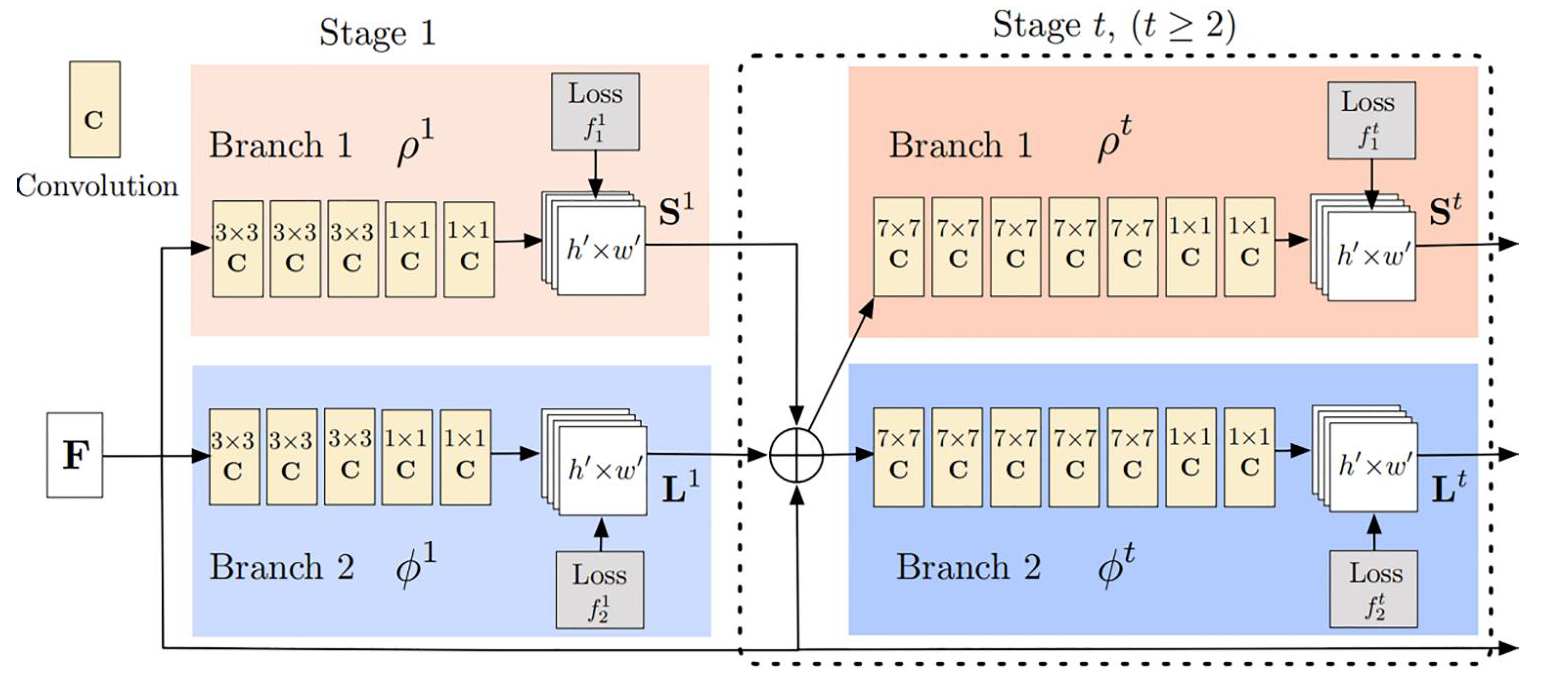}
	\caption{The architecture of the OpenPose model \cite{cao2017realtime}} 
	\label{fig6}
\end{figure*}
\subsubsection{Panoptic dataset}
Ground truth for these datasets was made with a motion capture system containing 480 VHA camera views, 31 HD views, 10 RGB-D sensors. It consists of 65 sequences with 5.5 total number of hours and more than 1.5 million 3D skeletons.
\subsubsection{3DPW dataset}
Ground truth for this dataset was made with a single hand-held camera. There are 60 video sequences with more than 51,000 frames. Figure \ref{fig4} shows some of the human poses from both datasets.
\begin{table}
	\footnotesize
	\centering
	\caption{Overview of 2D datasets}
	\label{tab1}
	\begin{tabular}{|l|l|l|l|l|}
		\hline
		Dataset &  Joints & Train & Val & Test \\ \hline
		%& & 2D datasets & & \\ \hline
		MPII &  16 & 3.8k  & 0 & 1.7k \\ \hline
		COCO &  17 & 64k  & 2k & 40k  \\ \hline
	\end{tabular}
\end{table}

\subsection{Performance metrics}
Performance metrics are important tools for evaluating the given method. There a couple of metrics that are specific for HPE problems and they are divided into 2D and 3D performance metrics.

\subsubsection{2D metrics}

\paragraph{Percentage of correct parts (PCP)\cite{ferrari2008progressive}}
It characterizes the accuracy of a limb’s predicted location. A limb is said to be accurately localized if two predicted keypoints that make the respective limb are within some threshold of the limb’s ground-truth keypoints. This threshold can be 50\% of the length of the predicted limb. PCP can either be represented as the mean value of all predictions, or it can be produced for a specific limb. PCPm \cite{andriluka20142d} is a variation of PCP where the threshold is represented as 50\% of the mean ground-truth segment length over the whole test set.
\paragraph{Percentage of correct keypoints (PCK)\cite{yang2012articulated}}
measures how accurate is predicted keypoint’s location. Keypoint location prediction is said to be correct if it’s in the given threshold pixels of the ground-truth keypoint. This threshold can be defined by the person’s bounding box, or most often as 50\% of the head segment length (PCKh@0.5) \cite{andriluka20142d}.
The Average Precision (AP) measures how accurate is the prediction, that is, what is the percentage of correct predictions. It takes into account true positive and false positive predictions and represents the ratio of true positive predictions and sum of true positive and false-positive predictions. For the evaluation of HPE methods if a predicted keypoint’s location falls within the threshold of the ground-truth keypoint location it is counted as a true positive. In the end, all the unassigned predictions are counted as false positive.
\paragraph{Average Precision (AP)}
It measures how accurate is the prediction, that is, what is the percentage of correct predictions. It takes into account true positive and false positive predictions and represents the ratio of true positive predictions and sum of true positive and false-positive predictions. For the evaluation of HPE methods if a predicted keypoint’s location falls within the threshold of the ground-truth keypoint location it is counted as a true positive. In the end, all the unassigned predictions are counted as false positive.

\subsubsection{3D metrics}
\paragraph{Mean Per Joint Position Error (MPJPE)}
It is a measure used for evaluating 3D HPE methods. This metric calculates Euclidean distance between the predicted 3D keypoint location and the ground-truth keypoint location. This is then averaged for all keypoints in an image, or, if there is a set of frames, the mean error is averaged for all frames [1].

\begin{table}
	\footnotesize
	\centering
	\caption{Overview of 3D datasets}
	\label{tab2}
	\begin{tabular}{|l|l|l|}
		\hline
		Dataset &  Joints & Train \\ \hline
		%3D datasets&&  \\ \hline
		Panoptic &  15 & 65 videos (5.5 hours)    \\ \hline
		3DPW     &  18 & 60 videos (51k frames)   \\		\hline
	\end{tabular}
\end{table}
\section{Overview of Popular algorithms}
\label{sec:4}
This section will give an overview of state-of-the-art methods both 2D and 3D.
\subsection{2D methods}

\subsubsection{OpenPose}
The detection-based method presented in [24] describes an approach that uses a non-parametric representation, called Part Affinity Fields (PAF). Some of the problems addressed in [24] are the possibility of detecting individual body joints even when in the given input there are multiple people, or there is irregular interaction between people, or the size of the people in the image is different. The workflow for this method is following. First, for a given input (Fig. \ref{fig5} a), a set of part confidence maps is predicted. These maps represent the location of each joint (Fig. \ref{fig5} b). Additionally, a set of part affinity fields is predicted (Fig. 5. c). These fields describe the location and orientation of the body parts and are described as a set of 2D vectors that indicate the degree of association between the body parts \cite{munea2020progress}. Afterward, a matching algorithm is needed to make an association between predicted joints (Fig. \ref{fig5} d). This is done with bipartite matching. In the end, these matched joints need to be assembled during the parsing phase (Fig. \ref{fig5} e) into a full-body pose for each person in the image. The whole workflow can be seen in Figure \ref{fig5}.

Figure \ref{fig6} shows the overall architecture of this method. This method is represented as multi-stage two-branch CNN architecture. In order to predict all the maps from a given input, the input image needs to be converted into feature space F. This is done with the help of the first 10 layers of the VGG network. The converted input is then fed into the two branches that work simultaneously. The beige branch in figure 6 represents the branch that is predicting set S of the part confidence maps with some confidence of J. The blue branch predicts a set of part affinity fields S of 2D vectors. There are C vectors corresponding to each body part. The first stage outputs a set of part confidence maps and part affinity fields. This is forwarded in all successive stages along with the feature map. Additionally, after each stage, there is intermediate supervision. This is done to solve the vanishing gradients.
\begin{figure*}[!htp]
	\centering
	\includegraphics[scale=0.6]{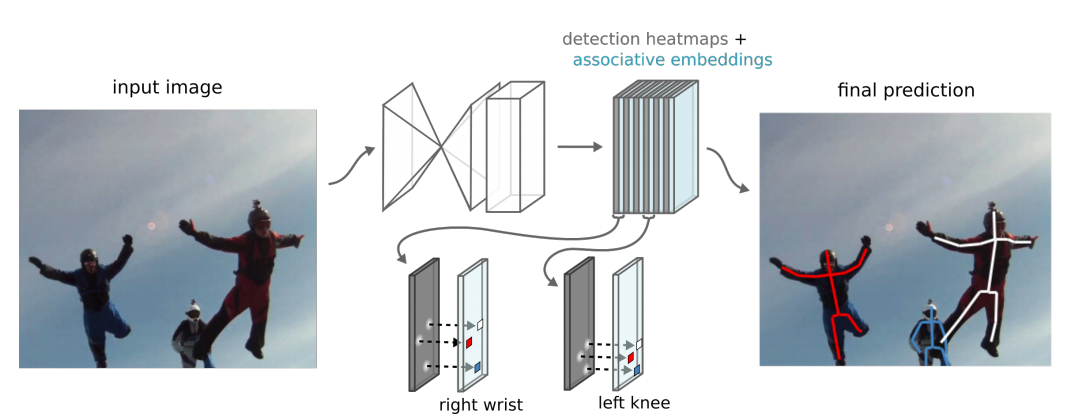}
	\caption{The architecture of associative embedding network for multi-person pose detection  \cite{newell2016associative}} 
	\label{fig7}
\end{figure*}
\subsubsection{Associative embedding}
The method that is presented in \cite{newell2016associative} instead of using multi-stages for human pose estimation, uses an end-to-end network. This is done by introducing the idea of associative embedding. Using this idea, it is possible to jointly perform detection and grouping with just one stage. The concept behind the idea is the following. For each detection, a vector embedding is introduced. This embedding is a tag number that is used to identify the group assignment. That is, all the detections that have the same tag value belong to the same body pose. The network outputs two heatmaps. A heatmap for per-pixel detection scores, which represents a map of detection scores at each pixel for each joint. The other heatmap is a heatmap for per-pixel identity tags. That is, tagging score at each pixel for each joint. These two heatmaps are then used together to extract the corresponding embedding from the pixel location with the highest detection score for each body joint. Two loss functions are combined. First is detection loss which is the mean square error between each predicted detection heatmap and its ground truth heatmap. The second is grouping loss which compares the tags within each person and across the people. Tags within a person should be the same, whereas tags for different people should be separate. It's worth noting that there is no ground truth for tags as what matters is not the exact value of the tag, but that there is some difference between tags of body joints of different people. In the case of multi-person pose estimation, associative embedding is integrated with a stacked hourglass network. This integrated architecture produces output that is represented as a detection heatmap and a tagging heatmap for each body joint. The outcome is then used to group body joints with similar tags into individual people's poses. The architecture for multi-person pose estimation is shown in figure \ref{fig7}. 

\subsection{3D methods}
This section gives an overview of 2 different state-of-the-art 3D bottom-up multi-person HPE methods.
\subsubsection{XNECT}
\begin{figure*}[!htp]
	\centering
	\includegraphics[scale=0.4]{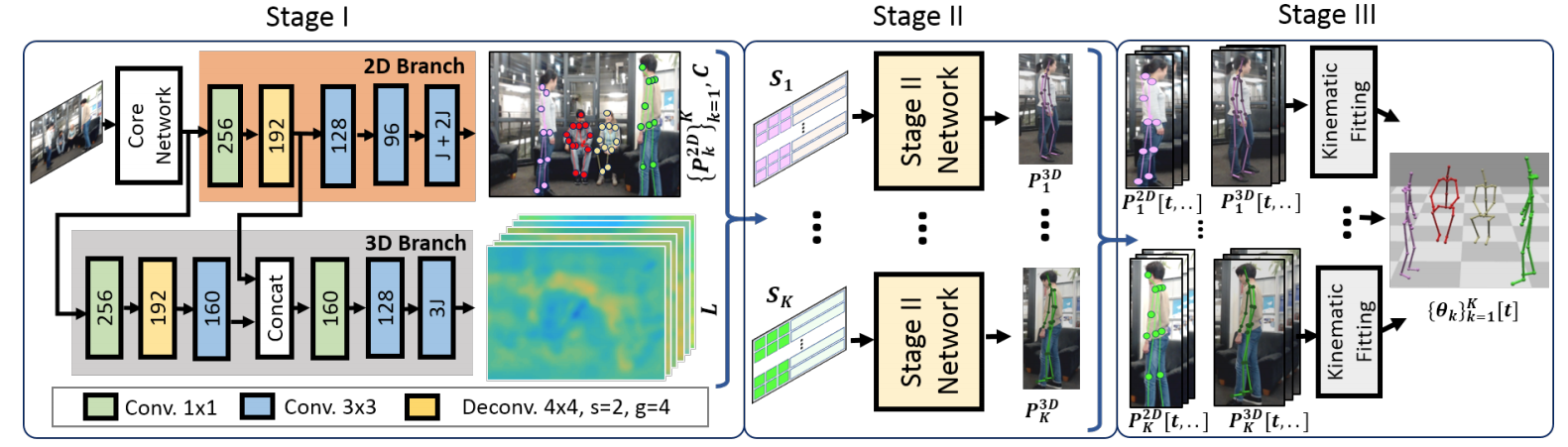}
	\caption{The overview of the architecture of XNect model \cite{mehta2020xnect}} 
	\label{fig8}
\end{figure*}

In the \cite{mehta2020xnect} method named XNect is presented. This method is a real-time method for 3D multi-person pose estimation from RGB image input. It is tailored to withstand generic scenes and occlusions by objects and people. The main idea behind it is to introduce two deep neural network stages that can have a local scope – for each body joint – and global scope – for all body joints. It is a bottom-up approach that is multi-staged. It contains three stages, where the last uses a predefined 3D skeleton to enhance the performance. Therefore, this method is model-based. Figure \ref{fig8} shows the architecture of this network. The first stage is a convolutional neural network that tries to estimate 2D and 3D pose features with the identity assignments. The network that is used is SelecSLSNet. This CNN contains selective long and short skip connections that improve the information flow and optimize the inference speed of the network. The output from this stage is a set of 2D part confidence maps that contain the information for assigning body parts to an individual. Besides this, the output of this stage is also an intermediate 3D pose encoding for the bones that connect at the joint. The key insight of this stage is that it only estimates the pose features of body joints that are not occluded in the image. Additional refinement is done in stage 2. This stage is comprised of a fully connected neural network. This stage estimates a full 3D pose regarding the possible partial 2D and 3D pose features for each person from the previous stage. The final third stage uses a skeletal model to optimize the 3D pose estimation. The space-time skeletal model predicts the 3D pose and enforces temporal coherence.

\subsubsection{MUBYNET}
\begin{figure}[!htp]
	\centering
	\includegraphics[scale=0.29]{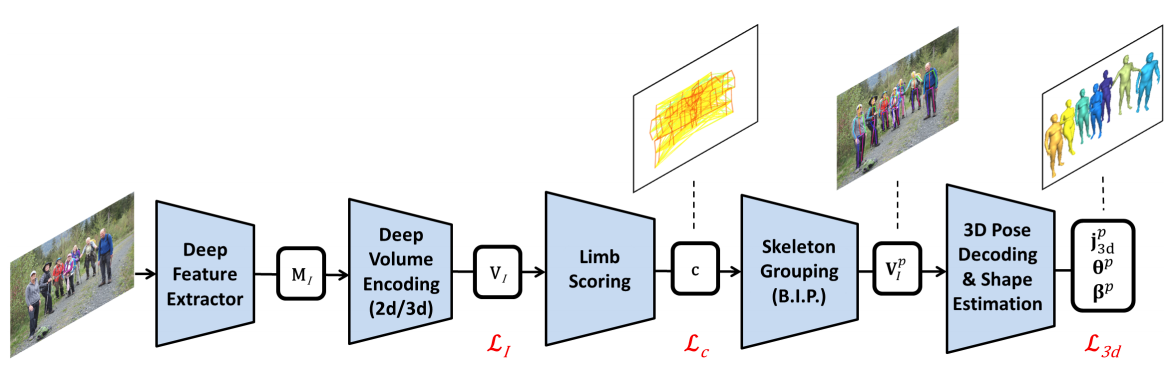}
	\caption{The overview of the architecture for the MubyNet model \cite{zanfir2018deep}} 
	\label{fig9}
\end{figure}

MubyNet method is presented in \cite{zanfir2018deep}. This is a multi-stage, not a real-time method used for 3D pose and shape estimation and integrated localization. It is a bottom-up method that can estimate poses for multiple people in the given input. Firstly, this method identifies body joints and limbs and groups them. This grouping is done with 2D and 3D information that is obtained with learned scoring functions. In the end, this method optimally aggregates estimated joints and limbs into a partial or complete 3D human body pose with the help of the skeleton hypothesis and kinematic tree constraints. Figure \ref{fig9} shows the overall architecture of MubyNet. The input in the network is an RGB image that goes into the deep feature extractor stage. This stage computes features that are used in the deep volume encoding stage. At this point, features are refined to contain only 2D and 3D pose information. The next stage is the limb scoring stage, where all possible kinematic connections between 2D detected joints are collected. Such collected links are then used to predict corresponding scores. These scores are then fed into the skeleton grouping stage, where people are localized by assembling limbs into skeletons. In the end, for each person in the 3D pose decoding \& shape estimation stage, 3D pose is estimated.
\begin{table}
	\footnotesize
	\centering
	\caption{Overview and evaluation of some methods for multi-person pose estimation }
	\label{tab3}
	\begin{tabular}{|p{1.6cm}|p{0.4cm}|p{.9cm}|p{.7cm}|p{.7cm}|p{.6cm}|}
		\hline
		Algorithm &  Type & Highlight & Dataset & Metrics& Score \\ \hline
		OpenPose &  16 & 3.8k  & 0 & 1.7k& Test \\  \hline
		Associative embeddling &  17 & 64k  & 2k & 40k  & Test\\ \hline
		Multi-Posenet &  17 & 64k  & 2k & 40k  & Test\\ \hline
		HRNet &  17 & 64k  & 2k & 40k  & Test\\ \hline
		PoseFix &  17 & 64k  & 2k & 40k  & Test\\ \hline
	\end{tabular}
\end{table}

\section{Summary and discussion}
\label{sec:5}
In the previous section, some of the most well-known bottom-up methods for multi-person 2D or 3D pose estimation were briefly explained. The reason for choosing these methods as an example lies in their innovation, influence, contribution, and different types of ideas and approaches that can be valuable to the reader. In this section, all previously explained methods are compared and discussed. The rest of the research paper will address the existing issues, but it will also give an insight into possible future directions for this type of research. Table 2 shows a short description of each method, its category, and comparable metrics. To have a better overview, the main requirement for a method shown in this table is to be a multi-person pose estimation method.
\begin{figure*}[!htp]
	\centering
	\subfloat[\centering]{{\includegraphics[width=7.5cm]{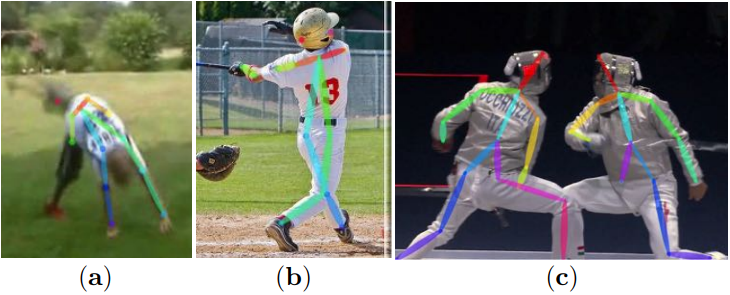} }}%
	\qquad
	\subfloat[\centering]{{\includegraphics[width=7.5cm]{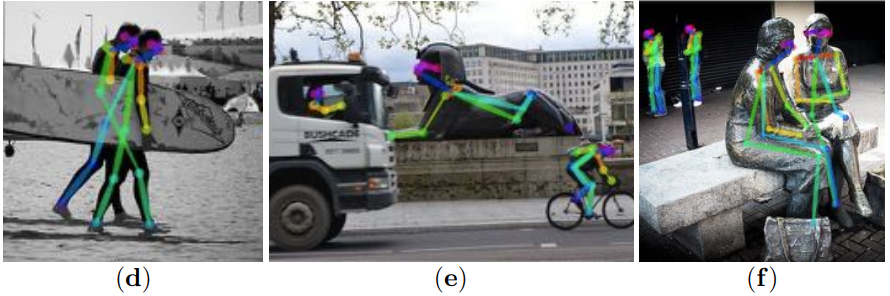} }}%
	\caption{(a) rare pose, (b) missing body part, (c) overlapping of different body parts, (d) assigning body part to a wrong person, (e-f) false positives: animals, statues, etc. \cite{cao2017realtime}}%
	\label{fig10}%
\end{figure*}
\subsection{Existing Issues}
\begin{figure}
	\centering
	\includegraphics[scale=0.5]{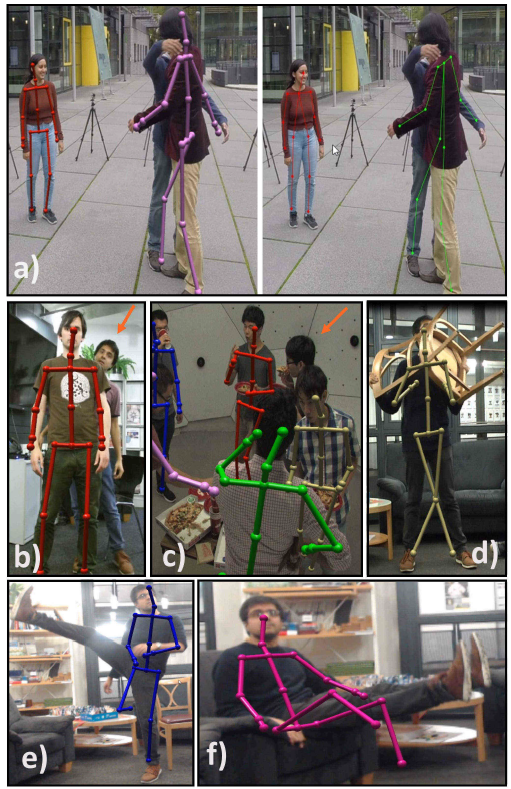}
	\caption{a) 2D pose limb confusion, (b-c) person not detected due to neck occlusion, (d) body orientation confusion due to occluded face, (e-f) poses not seen in the training set \cite{mehta2020xnect}}
	\label{fig11}
\end{figure}

In order to have a complete picture of bottom-up methods, some of the existing issues need to be addressed. From table 2, it can be observed that top-down approaches yield better results metrics-wise. But for a method to be used in real-life, the inference time is often more important than the other metrics values. Bottom-up approaches still show sufficiently better speed-wise results and are worth looking into. However, it is essential to note that the performance of bottom-up methods heavily depends on how complex the background is and the frequency of human occlusions.  Some of these failure cases can be seen in figure \ref{fig10}.
Specifically, for the 3D bottom-up approaches for multi-person pose estimation, the following issues arise.  For example, the MubyNet method encodes the full 3D pose vector for each pixel part of the skeleton instead of just encoding it at the body joints' location. This is a suboptimal solution, and it creates potential encoding conflicts in the 3D feature space. On the other hand, XNect uses a different approach. This method tries to keep it simple in the beginning and encodes the 3D pose vector just at the body joint location, but in later stages, it fills out the missing information for the rest of the pixels. This makes the problem more manageable and robust to wrongly encoded 3D pose vectors. XNect method doesn’t give good results when 2D pose estimates that are used inside of the network are wrong. Additionally, as it was stated in the XNect paper, it is required for the neck to be visible for successful detection. If this is not the case, XNect cannot give a good detection and estimation. The rest of the limitations can be seen in figure \ref{fig11}. 

\subsection{Future directions}
In order to resolve some of the current issues of HPE, future work will need to look into both global and local contexts when estimating the human pose. Some of the confusion could potentially be resolved with cues from both contexts. Besides this, future networks need to be more efficient. How fast is the inference mode for a particular method significantly impacts how and when these methods will be used in real-life. Additionally, because of how valuable datasets are in these kinds of problems, the datasets could also be diversified. That way, the methods can be more robust and handle complex scenes. The other strategy is to create datasets for just a specific complex scenario where the proposed method will be used. For the 3D multi-person pose estimation methods, some of the physics-based motion constraints can be employed. These constraints could lead to the improvement of pose stability and overall temporal stability. Furthermore, these problems could be remedied with the help of a better space-time identity tracker.
\begin{table}
	\footnotesize
	\centering
	\caption{Overview and evaluation of some methods for multi-person pose estimation }
	\label{tab4}
	\begin{tabular}{|p{1.6cm}|p{0.4cm}|p{.9cm}|p{.7cm}|p{.7cm}|p{.6cm}|}
		\hline
		Algorithm &  Type & Highlight & Dataset & Metrics& Score \\ \hline
		XNect &  16 & 3.8k  & 0 & 1.7k& Test \\  \hline
		MubyNet &  17 & 64k  & 2k & 40k  & Test\\ \hline
		LCR-Net++ &  17 & 64k  & 2k & 40k  & Test\\ \hline
		Camera distance aware &  17 & 64k  & 2k & 40k  & Test\\ \hline
	\end{tabular}
\end{table}
\section{Conclusion:}
\label{sec:6}
This paper represents an overview of the carefully selected 2D and 3D bottom-up multi-person pose estimation methods. These methods are promising and necessary for future applications in real-life problems. It is worth noting that the bottom-up multi-person pose estimation methods give a real-time performance. Although bottom-up approaches are in some cases worse than top-down approaches, bottom-up approaches still hold the top of the list when the inference mode speed is in the case. This review paper briefly introduces the human pose estimation research field, and it starts with problem formulation and introduction to human pose estimation.
Furthermore, the overview of the most commonly used datasets is given, alongside the primarily used evaluation metrics in this field. Additionally, an overview of some of the prominent methods is given, and finally, these methods alongside other state-of-the-art methods are compared and discussed. To sum up, this review paper provides the reader with an essential background of this topic. Additionally, it presents an overview of some of the limitations and open issues that can help future researchers wisely choose the architecture and be aware of all the caveats and potential future research directions.

%------------------------------------------------------------------------------------------------------------------%

%\section*{Acknowledgments}
%This research was funded by Norwegian Ministry of education and research and Hainan provincial scientific collaboration project under grant KJHZ2015-23.
%------------------------------------------------------------------------------------------------------------------%

{\small
\bibliographystyle{IEEEbib}
\bibliography{Milan_Survey}
%\bibliography{strings,refs,external}
}
\end{document}